\documentclass{article}

\usepackage{microtype}
\usepackage{graphicx}
\usepackage{subfigure}
\usepackage{booktabs} 
\usepackage{hyperref}

\usepackage[accepted]{icml2024}
\usepackage{amsmath}
\usepackage{amssymb}
\usepackage{mathtools}
\usepackage{amsthm}
\usepackage{multicol}

\usepackage[capitalize,noabbrev]{cleveref}

\theoremstyle{plain}
\newtheorem{theorem}{Theorem}[section]

\newtheorem{lemma}[theorem]{Lemma}

\theoremstyle{definition}

\newtheorem{assumption}[theorem]{Assumption}
\theoremstyle{remark}

\usepackage[textsize=tiny]{todonotes}

\usepackage{bbm}
\usepackage{wrapfig}
\usepackage[utf8]{inputenc} 
\usepackage[T1]{fontenc}    
\usepackage{hyperref}       
\usepackage{url}            
\usepackage{booktabs}       
\usepackage{amsfonts}       
\usepackage{nicefrac}       
\usepackage{microtype}      
\usepackage{xcolor}         
\usepackage[utf8]{inputenc} 
\usepackage[T1]{fontenc}    
\usepackage{url}            
\usepackage{booktabs}       
\usepackage{amsfonts}       
\usepackage{nicefrac}       
\usepackage{lipsum}		
\usepackage{graphicx}
\usepackage{natbib}
\usepackage{doi}
\usepackage{caption}
\usepackage[american]{babel}
\usepackage{subfigure}
\usepackage{booktabs} 
\usepackage[normalem]{ulem}
\usepackage{tikz}
\usetikzlibrary{arrows.meta, automata,
                positioning,
                quotes}
\usepackage{pst-node}
\usepackage{wrapfig}
\usepackage{lipsum}
\usepackage{verbatim}
\allowdisplaybreaks
\usepackage{microtype}
\usepackage{color}
\usepackage{eucal}
\usepackage{epsfig,amsmath,amssymb,amstext,amsthm,mathrsfs}
\usepackage{graphics,epsf,epsfig,psfrag}
\usepackage{dsfont,color,epstopdf,fixmath}
\usepackage{enumitem}
\usepackage{algorithm,algorithmic,latexsym}

\usepackage[acronym]{glossaries} 

\newacronym{T1}{T1}{\theta_1}
\usepackage{color}

\newcommand{\mcs}{\mathcal{S}}
\newcommand{\mca}{\mathcal{A}}
\newcommand{\nn}{\nonumber}

\newcommand{\cp}{\mathcal{P}}

\usepackage{mathtools} 
\usepackage{siunitx} 
\usepackage{booktabs} 
\usepackage{tikz} 
\usepackage{cleveref}
\usepackage[toc,page,header]{appendix}
\usepackage{minitoc}

\DeclareMathOperator*{\argmax}{arg\,max}

\begin{document}

\twocolumn[
\icmltitle{Constrained Reinforcement Learning Under Model Mismatch}


\begin{icmlauthorlist}
\icmlauthor{Zhongchang Sun}{yyy}
\icmlauthor{Sihong He}{comp}
\icmlauthor{Fei Miao}{comp}
\icmlauthor{Shaofeng Zou}{yyy}
\end{icmlauthorlist}

\icmlaffiliation{yyy}{Department of Electrical Engineering, University at Buffalo, New York, USA}
\icmlaffiliation{comp}{Department of Computer Science and Engineering, University of Connecticut, Storrs, USA}

\icmlcorrespondingauthor{Shaofeng Zou}{szou3@buffalo.edu}

\icmlkeywords{Machine Learning, ICML}

\vskip 0.3in
]

\printAffiliationsAndNotice{}

\begin{abstract}
Existing studies on constrained reinforcement learning (RL) may obtain a well-performing policy in the training environment.  However, when deployed in a real environment, it may easily violate constraints that were originally satisfied during training because there might be model mismatch between the training and real environments. 
To address the above challenge, we formulate the problem as constrained RL under model uncertainty, where the goal is to learn a good policy that optimizes the reward and at the same time satisfy the constraint under model mismatch. We develop a Robust Constrained Policy Optimization (RCPO) algorithm, which is the first algorithm that applies to large/continuous state space and has theoretical guarantees on worst-case reward improvement and constraint violation at each iteration during the training. We demonstrate the effectiveness of our algorithm on a set of RL tasks with constraints.

\end{abstract}

\section{Introduction}
\label{sec:intro}
In reinforcement learning (RL), the agent aims to learn a policy that maximizes the expected cumulative reward by interacting with an environment \cite{sutton2018reinforcement}. However, in real-life applications, e.g.,  robotics \cite{levine2016end,ono2015chance}, health care \cite{yu2019reinforcement},  autonomous driving \cite{kiran2020deep,fisac2018general} and industry automation \cite{gasparik2018safety}, where it is crucial to meet various constraints while maximizing reward,  application of RL still remains limited. For example,  an unmanned aerial vehicle (UAV) performing post-disaster search and rescue needs to return and charge before running out of battery, and communication system needs to maximize throughput while adhering to power consumption and latency constraints. 


The framework of constrained Markov Decision Process (CMDP)  was developed \cite{altman1999constrained} to tackle the above challenge and the  goal is to search for one  policy that maximizes the overall reward among the policies that satisfy the constraint, and an optimal policy for CMDP can be found via linear programming.

When a well-performing policy trained using a simulator is deployed in a real environment, it may easily violate constraints that were originally satisfied during training because there might be model mismatch between the training and real environments. This could be due to environment non-stationarity, sim-to-real gap and adversarial attacks. Despite its practical importance, studies on the problem of robust RL under constraints are rather limited in the literature. 
Several attempts were made in \cite{russel2020robust,mankowitz2020robust}, where two heuristic approaches were proposed. Their basic idea  is to first evaluate the worst-case performance of the policy over the uncertainty set, and then use that together with classical policy improvement methods, e.g., policy gradient, to update the policy. However, there is no guarantee to obtain an improved robust policy  by doing so. A robust primal-dual approach was developed in \cite{wang2022robust}, which however cannot guarantee monotonic robust reward improvement or constraint satisfaction during the training. Also, the results in \cite{wang2022robust} are limited to the tabular case with finite state and action spaces.


In this paper, we study the problem of constrained RL under model mismatch. Specifically, we consider an uncertainty set of transition kernels
that characterizes the potential model mismatch (see \cite{siddiquerobust} for an example of uncertainty set construction). The goal is to guarantee that for any MDP in the uncertainty set, the constraint is always satisfied. Among these policies, we aim to identify one that maximizes the worst-case accumulated reward over the uncertainty set. 
Solution to the above problem is robust in that for any MDP in the uncertainty set, the constraint is always satisfied, and at the same time, the overall reward of the obtained policy is also robust to model mismatch.

We develop a robust constrained policy optimization (RCPO) algorithm, and theoretically bound the constraint violation for any transition kernel in the uncertainty set
, and the worst-case reward improvement over the uncertainty set for every policy during training. One thing to highlight is that our algorithm applies to Markov Decision Processes (MDPs) with continuous state space, which allows applications to large scale practical problems.

One essential theoretical result that drives our RCPO algorithm development is a generalization of the performance difference lemma \cite{kakade2002approximately,achiam2017constrained} to robust MDPs. Specifically, we consider the robust value function, which measures the worst-case performance of a policy over the uncertainty set. We bound the difference between robust value functions for two different policies using the difference between the two policies. 



Our algorithm consists of two steps for each update: (i) robust policy improvement step and (ii) projection. 

Step (i) uses our robust performance difference lemma to develop a local approximation of the robust value function, and design a robust policy improvement step that searches in the neighborhood of the current policy. This step generalizes the trust region method \cite{schulman2015trust} to the robust setting and guarantees robust reward improvement. 
Unlike the non-robust setting where there is only one transition kernel which  stays the same throughout the training, under model uncertainty the worst-case transition kernel changes with the policy and is different for reward and utility (see \Cref{sec:rmdp}). One challenge is that the local approximation implicitly depends on the policy to be optimized, which is in the neighborhood of the current policy, through its worst-case transition kernel, making the optimization intractable. We develop a novel approximation using the current policy as a surrogate, and prove that such an approximation still provides guaranteed robust reward improvement (and later in step (ii) robust constraint satisfaction).


The obtained policy guarantees the reward improvement, but may violate the constraint due to bad initialization and stochastic noise. This leads to a potential problem of the constrained policy optimization (CPO) approach in \cite{achiam2017constrained} that there may not exist any feasible solution during the updates (as pointed out in \cite{yang2019projection}).  To address this challenge, in step (ii) we further project the obtained policy so that it satisfies the constraint for any transition kernel in the uncertainty set. 

One of the key challenges in the analysis lies in that the worst-case transition kernel changes when the policy updates. We address this challenge by leveraging our robust performance difference lemma and a novel integration of the Lipchitz property of robust value function and the change of measure technique.




\section{Related Work}

\textbf{Constrained RL.}
The CMDPs \cite{altman1999constrained} have been an active field of research. The primal-dual method is one of the most commonly used method \cite{altman1999constrained, auer2008near, liang2018accelerated, paternain2019constrained, tessler2018reward, yu2019convergent, stooke2020responsive, efroni2020exploration, zheng2020constrained, zhang2020first}, which converts the constrained optimization problem to an unconstrained Lagrangian formulation and alternatively updates the primal and dual variables. Thanks to the strong duality of the non-robust CMDPs \cite{paternain2019constrained}, the problem can be solved exactly in the dual domain and the primal-dual method is guaranteed to converge to the optimal solution \cite{ding2020natural, ding2021provably, li2021faster, liu2021fast, ying2021dual, wei2022triple-q}. Another widely studied method is the primal method \cite{chow2018lyapunov, dalal2018safe, liu2020ipo, xu2021crpo, bura2022dope}, which takes all the updates in the primal domain without formulating the Lagrangian function. The trust region-based methods have also been proposed to solve the non-robust CMDPs \cite{achiam2017constrained, yang2019projection, kim2023trust}, which guarantee the reward improvement and constraint satisfaction during the training. Under model mismatch,  the worst-case transition kernel is different as the policy updates, and therefore these approaches may not be applied, and the obtained policy may easily violate the constraint when there is a  model mismatch. 

\textbf{Robust RL.} Robust RL was firstly introduced in \cite{iyengar2005robust, nilim2004robustness} where the goal is to optimize the worst-case performance over the uncertainty set of transition kernels. Algorithms with convergence guarantee have been proposed for both the model-based robust RL with known uncertainty set \cite{iyengar2005robust, nilim2004robustness, xu2010distributionally, lim2019kernel, wang2023policy,wang2023robust} and the model-free robust RL with unknown uncertainty set \cite{roy2017reinforcement,  zhou2021finite, panaganti2021sample, yang2021towards, wang2022policy,wang2023model,wang2021online,wang2022robust}. Compared with the unconstrained robust RL, the robust constrained RL is more challenging since we also need to guarantee the constraint is satisfied for any transition kernel in the uncertainty set. Directly applying algorithms designed for unconstrained robust RL to robust constrained RL will lead to constraint-violating policies.
 
\textbf{Constrained RL under Model Uncertainty.}
Unlike the non-robust CMDPs, there are rather limited works on constrained RL under model uncertainty. In \cite{russel2020robust}, a heuristic approach is proposed. The basic idea is that they first estimate the robust value function, and then update the policy using the non-robust policy gradient \cite{sutton1999policy}. Since the worst-case transition kernel is also a function of the policy, the non-robust policy gradient may not update the policy along the direction of the real gradient. Therefore, it cannot guarantee the performance improvement for each update, and thus the convergence of this heuristic approach may not hold. In \cite{mankowitz2020robust}, the robust value function estimate is first performed and a non-robust continuous control algorithm is applied to update the policy. Similar to \cite{russel2020robust}, the non-robust policy improvement cannot guarantee the convergence of the algorithm. In \cite{wang2022robust}, a robust primal-dual algorithm (RPD) is proposed where the primal and dual variables are updated alternatively, and the robust policy gradient is employed to update the policy. However, the strong duality may not hold when there is model mismatch. Secondly, the constraint may be violated during the training, which is attributed to the nature of the primal-dual approach. In contrast, the update of our algorithm is performed in the primal domain, and we provide performance guarantee on the robust reward improvement and robust constraint violation for each update. Our RCPO ensures constraint satisfaction throughout the training, which is critical for many practical applications.

\section{Preliminaries and Problem Formulation}
\subsection{Constrained MDP}
A constrained Markov Decision Process (CMDP) \cite{altman1999constrained} is defined by a tuple $(\mcs, \mca, p, r, c)$, where $\mcs$ is the state space, $\mca$ is the action space, $p = \{p^a_s\in\Delta(\mcs), s\in \mcs, a\in\mca\}$\footnote{$\Delta(\mcs)$ denotes the probability simplex defined on $\mcs$.} is the transition kernel of the CMDP, $r: \mcs \times \mca \rightarrow [0, 1]$ is the reward function, and $c: \mcs \times \mca \rightarrow [0, 1]$ is the utility function. A stationary policy $\pi: \mcs \rightarrow \Delta(\mca)$ is defined as the probability distribution of choosing actions in $\mca$ at the current state $s$. After choosing action $a$ at state $s$, the system transits to the next state $s^\prime$ based on the transition kernel $p^a_s$. At the same time, the agent receives a reward $r(s, a)$ and a utility $c(s, a)$. For the sake of simplicity in presentation, we consider the case with one constraint, and the results in this paper can be extended to the case with multiple constraints.

Starting from an initial state $s$, the reward value function of a policy $\pi$ is defined as 
\begin{flalign*}
   V_{r,p}^\pi(s) \triangleq \mathbb{E}_{p}\left[\sum_{t=0}^\infty \gamma^t r(s_t, a_t)|s_0 = s, \pi\right], 
\end{flalign*}
where $\mathbb{E}_{p}$ denotes the expectation with respect to the transition kernel $p$ and $\gamma$ is the discount factor. The reward action value function is defined as
\begin{flalign*}
    Q_{r,p}^\pi(s, a) \triangleq \mathbb{E}_{p}\left[\sum_{t=0}^\infty \gamma^t r(s_t, a_t)|s_0 = s, a_0 = a, \pi\right].
\end{flalign*}
Similarly, the utility value function and the utility action value function are defined as 
\begin{flalign*}
    &V_{c, p}^\pi(s) \triangleq \mathbb{E}_{p}\left[\sum_{t=0}^\infty \gamma^t c(s_t, a_t)|s_0 = s, \pi\right]\nn\\&
    Q_{c, p}^\pi(s, a) \triangleq \mathbb{E}_{p}\left[\sum_{t=0}^\infty \gamma^t c(s_t, a_t)|s_0 = s, a_0 = a, \pi\right].
\end{flalign*} 
Let $V_{r,p}^\pi(\rho) = \mathbb{E}_{s\sim\rho}[V_{r, p}^\pi(s)]$ and $V_{c,p}^\pi(\rho) = \mathbb{E}_{s\sim\rho}[V_{c, p}^\pi(s)]$ be the discounted accumulative reward function and the discounted accumulative utility function, respectively, when the initial state $s$ follows distribution $\rho$. Let $d^{\pi}_p(s)$ denote the state occupancy measure when the initial state $s$ follows distribution $\rho$: 
\begin{flalign*}
&d^{\pi}_p(s) = (1-\gamma)\sum_{t=0}^\infty \gamma^t \mathbb{P}(s_t = s|s_0\sim\rho,\pi, p).
\end{flalign*}
The goal of the CMDP is to learn a policy $\pi$ that maximizes the cumulative discounted reward $V_{r,p}^\pi(\rho)$ subject to the constraint on the cumulative discounted utility $V_{c,p}^\pi(\rho)$, i.e.,
\begin{flalign*}
    \max_\pi V_{r,p}^\pi(\rho)\ \text{s.t.}\ V_{c,p}^\pi(\rho) \geq d,
\end{flalign*}
where $d$ is some positive threshold for the constraint.

\subsection{Robust MDP}\label{sec:rmdp}
The robust MDP is defined as $(\mcs, \mca, \cp, r)$, where $\cp$ is the uncertainty set of transition kernels that measures the model uncertainty. In this paper, we consider the uncertainty set with $(s, a)$-rectangularity \cite{nilim2004robustness, iyengar2005robust}. Specifically, the uncertainty set is defined as $\cp = \bigotimes_{s, a} \cp_s^a$, where $\cp_s^a \subseteq \Delta(\mcs)$ are independent over different state-action pairs. The robust value function is defined as
\begin{flalign}\label{eq:robustr}
    V_r^\pi(s) \triangleq \min_{p\in\cp}\mathbb{E}_{p}\Big[\sum_{t=0}^\infty \gamma^t r(s_t, a_t)|s_0=s, \pi\Big].
\end{flalign}
Similarly, the robust action value function is defined as 
\begin{flalign}\label{eq:robustq}
    Q_r^\pi(s, a) \triangleq \min_{p\in\cp}\mathbb{E}_{p}\Big[\sum_{t=0}^\infty \gamma^t r(s_t, a_t)|s_0 = s, a_0 = a, \pi\Big].
\end{flalign}
The transition kernel that achieves the min in \eqref{eq:robustr} and \eqref{eq:robustq} is referred to as the worst-case transition kernel. 
Denote by $V_{r}^\pi(\rho)$ the robust discounted accumulative reward function when the initial state $s$ follows the distribution $\rho$. For robust RL, the goal is to find an optimal robust policy $\pi^*$ that optimizes the worst-case performance over the uncertainty set of transition kernels, i.e.
\begin{flalign}\label{eq:goal}
\pi^* = \argmax_{\pi} V_r^\pi(\rho).
\end{flalign}

\subsection{Problem Formulation}
Define the constrained MDP problem under model mismatch as a tuple $(\mcs, \mca, \cp, r, c)$, where $\cp$ is an uncertainty set of transition kernels as defined in \Cref{sec:rmdp} to characterize the potential model mismatch (see \cite{siddiquerobust} for an example of uncertainty set construction). To guarantee that the constraint is always satisfied even under model mismatch, we define the robust utility value function which measures the worst-case accumulated utility over the uncertainty set:
\begin{flalign}\label{robustc}
V_c^\pi(s) \triangleq \min_{p\in\cp}\mathbb{E}_{p}\Big[\sum_{t=0}^\infty \gamma^t c(s_t, a_t)|s_0=s, \pi\Big].
\end{flalign}
We are interested in policies that for any transition kernel in the uncertainty set, i.e., under model mismatch, the accumulative utility is still above a prescribed threshold. Furthermore, among those policies, we would like to find one that achieves a good accumulative reward for any transition kernel in the uncertainty set.
Formally, we aim to find a policy that maximizes the worst-case cumulative discounted reward subject to the constraint on the worst-case cumulative discounted utility, i.e., 
\begin{flalign}\label{eq:problem}
\max_{\pi} V_r^\pi(\rho),\ \text{s.t.}\ V_c^\pi(\rho) \geq d.
\end{flalign}
The problem in \eqref{eq:problem} is referred to as robust constrained RL in this paper.

\section{Robust Constrained Policy Optimization}
In this section, we present our algorithm, the robust constrained policy optimization (RCPO), to solve the problem in \eqref{eq:problem}, and theoretically prove that the obtained policy has an improved robust reward value function  and also has guarantees for constraint satisfaction at each iteration. Our RCPO algorithm can be applied to large scale problems with a continuous state space. We also generalize the performance difference lemma in \cite{achiam2017constrained} to the robust setting, and show the robust value functions of two policies can be bounded using the divergence between them.

In this section, we first present our algorithm  and its theoretical performance analyses. In \Cref{sec:practical}, we will provide a practical implementation for an efficiently computation.

\subsection{Algorithm Design}

In the following, we will develop our RCPO algorithm. The basic idea is to first find a policy to maximize the robust reward advantage function in a neighborhood of the current policy, which generalizes the trust region policy optimization \cite{schulman2015trust} to the robust constrained RL problem, and then to project the obtained policy to meet the robust constraint. 

To obtain a local approximation of the robust value function, we first present the robust performance difference lemma. 
Specifically, we need a bound for the performance difference of the robust value functions between two policies. Let $p_{\pi}^{r}$ denote the worst-case transition kernel of $\pi$ for reward such that 
$V_{r,p_{\pi}^{r}}^\pi(\rho) = \min_{p\in\cp}V_{r,p}^\pi(\rho)$. Let $D_{KL}(f_0\|f_1)$ denote the Kullback-Leibler (KL) divergence between two distributions $f_0$ and $f_1$. 
The following robust performance difference lemma  generalizes the bound for standard non-robust value functions in \eqref{eq:per_diff} \cite{achiam2017constrained} to the robust value functions.
\begin{lemma}[Robust performance difference lemma]\label{lemma:rob_perf}
For any two policies $\pi, \pi^\prime$, let $$\epsilon^{\pi^\prime}_{r, p_{\pi^\prime}^{r}}=\max_s |\mathbb{E}_{a\sim\pi^\prime}[A^\pi_{r, p_{\pi^\prime}^{r}}(s, a)]|.$$ We have the following bound:
\begin{small}
\begin{flalign}\label{eq:lemma31}
    &V_r^{\pi^\prime}(\rho) - V_r^\pi(\rho)\geq\\
    &\frac{1}{1-\gamma} \mathbb{E}_{\substack{s\sim d^{\pi}_{p_{\pi^\prime}^{r}}\\a\sim \pi^\prime}}\left[A^\pi_{r, {p_{\pi^\prime}^{r}}}(s, a) -\frac{2\gamma\epsilon^{\pi^\prime}_{r, p_{\pi^\prime}^{r}}}{1-\gamma}\sqrt{\frac{1}{2}D_{KL}(\pi^\prime\|\pi)(s)}\right].\nn
\end{flalign} 
\end{small}
\end{lemma}
It can be easily verified that the bound in \eqref{eq:lemma31} holds with equality when $\pi = \pi^\prime$.

A first idea is to optimize the lower bound in \eqref{eq:lemma31} over $\pi^\prime$ in the neighborhood of $\pi$, and to obtain policy $\pi^\prime$ with an improved performance. However, it's difficult to implement since the lower bound in \eqref{eq:lemma31} involves the advantage function and visitation distribution under $p_{\pi^\prime}^{r}$ which implicitly depends on $\pi^\prime$. To address this unique challenge to the robust setting, we propose to approximate $A^\pi_{r, {p_{\pi^\prime}^{r}}}(s, a)$ and $d^{\pi}_{p_{\pi^\prime}^{r}}$ by $A^\pi_{r, {p_{\pi}^{r}}}(s, a)$ and $d^{\pi}_{p_{\pi}^{r}}$ respectively in the neighborhood of $\pi$. 
The motivation of such an approximation is that $\pi'$ is in the neighborhood of $\pi$, and the robust value function is Lipschitz in the policy (as shown in \cite{wang2023policy}).
We then use 
\begin{small}
\begin{flalign}\label{eq:lemma31app}
\frac{1}{1-\gamma} \mathbb{E}_{\substack{s\sim d^{\pi}_{p_{\pi}^{r}}\\ a\sim \pi^\prime} }\left[A^\pi_{r, {p_{\pi}^{r}}}(s, a)-\frac{2\gamma\epsilon^{\pi^\prime}_{r, p_{\pi^\prime}^{r}}}{1-\gamma}\sqrt{\frac{1}{2}D_{KL}(\pi^\prime\|\pi)(s)}\right]
\end{flalign}
\end{small}
as an approximation of the robust performance difference $V_r^{\pi^\prime}(\rho) - V_r^\pi(\rho)$ and further design our RCPO algorithm based on this approximation. 

As will be shown below, though \eqref{eq:lemma31app} may not necessarily be a lower bound of $V_r^{\pi^\prime}(\rho) - V_r^\pi(\rho)$ due to the use of the approximation, we are still able to guarantee both the reward improvement and the constraint violation. This actually corresponds to the additional challenge than the non-robust standard CMDP, where there is only one transition kernel for both policies. Here, we are interested in the robust value function, which is essentially the value function under the worst-case transition kernel, and two different policies induce two different worst-case transition kernels.

Let $p_k^r$ denote the worst-case transition kernel of $\pi_k$ for reward and $p_k^c$ denote the worst-case transition kernel of $\pi_k$ for utility. A direct generalization of the CPO algorithm in \cite{achiam2017constrained} is to optimize the policy iteratively using the following update:
\begin{flalign}\label{eq:rcpo}
    \pi_{k+1}&= \argmax_{\pi}\mathbb{E}_{\substack{s\sim d^{\pi_k}_{p_{k}^r}\\\hspace{-0.1cm}a\sim \pi}}\Big[A^{\pi_k}_{r, p_{k}^r}(s, a)\Big]\nn\\
    \text{s.t.}&\ V_{c, p_{k}^c}^{\pi_k}(\rho) +\frac{1}{1-\gamma} \mathbb{E}_{\substack{s\sim d^{\pi_k}_{p_{k}^c}\\\hspace{-0.1cm}a\sim \pi}}\Big[A^{\pi_k}_{c, p_{k}^c}(s, a)\Big] \geq d, \nn\\& \mathbb{E}_{s\sim d^{\pi_k}_{p_{k}^r}}[D_{KL}(\pi\|\pi_k)(s)]\leq \delta.
\end{flalign}
Here, the first constraint in \eqref{eq:rcpo} guarantees the new policy satisfies the robust constraint, and the second constraint in \eqref{eq:rcpo} limits the search to be in the neighborhood of $\pi_k$.
However, this has an issue that there may be no feasible solution to \eqref{eq:rcpo} if the current policy $\pi_k$ violates the constraint. 

To address the above challenge, we design a two-step approach which performs policy improvement followed by a projection step \cite{yang2019projection}. 
Below, we introduce our RCPO algorithm in details, and the pseudocode is provided in \cref{alg:rcpo}. To handle large-scale MDPs, we consider a parameterized policy class $\Pi_{\boldsymbol{\theta}}$ with parameter $\boldsymbol{\theta}$.

\textbf{Step 1: Robust Policy Improvement.} At the robust policy improvement step, we first estimate the worst-case transition kernel $p_k^{r}$ for the current policy $\pi_k$. This can be done by a gradient-based method \cite{wang2023policy}. We iteratively update $p_k^{r, t}$ using the projected gradient descent as follows,
\begin{flalign}
    p_k^{r, t+1} = \text{Proj}_{\cp}\big(p_k^{r, t} - \beta_t\nabla_{p}V_{r, p_k^{r, t}}^{\pi_k}(\rho)\big),
\end{flalign}
where $\beta_t$ is the step size and $\text{Proj}_{\cp}$ is the projection operator onto set $\cp$:
$
\text{Proj}_{\cp}(p_s^a) = \arg\min_{q\in\cp_s^a}D(p_s^a, q),
$
where $D$ is some distance measure between two distributions.

Consider the tabular case for an example, an accurate $p_k^{r}$ can be obtained such that
\begin{flalign}\label{eq:worst_r}
V_{r, p_k^{r}}^{\pi_k}(\rho) = \min_{p\in\cp}V_{r, p}^{\pi_k}(\rho),
\end{flalign}
as shown in Theorem 4.4 in \cite{wang2023policy}. 
For the large/continuous state space, to estimate the worst-case transition kernel, we parameterize the transition kernel and perform gradient descent to learn the worst-case transition kernel estimate. Consider the case with a large discrete state space as an example, the transition kernel can be parameterized as follows:
\begin{flalign}
    p_{s, a}^{\boldsymbol{\xi}}(s^\prime) = \frac{p^0_{s, a}(s^\prime)\cdot\exp(\frac{\boldsymbol{\eta}^\top\boldsymbol{\phi}(s^\prime)}{\lambda_{s, a}})}{\sum_{x}p^0_{s, a}(x)\exp(\frac{\boldsymbol{\eta}^\top\boldsymbol{\phi}(x)}{\lambda_{s, a}})},
\end{flalign}
where $p^0$ is the nominal transition kernel of the uncertainty set $\cp$, $\boldsymbol{\phi}: \mcs\rightarrow\mathbb{R}^m$ is a $m$-dimensional feature vector, $\boldsymbol{\xi} = (\boldsymbol{\eta}, \boldsymbol{\lambda})$, $\boldsymbol{\lambda} = \{\lambda_{s, a}>0, \forall (s, a)\in \mcs\times\mca\}$ and $\boldsymbol{\eta}\in\mathbb{R}^m$ are parameters. 
We then present another example for the case with a continuous state space, where the transition kernel can be parameterized using the Gaussian mixture model:
\begin{flalign}
    p_{s, a}^{\boldsymbol{\xi}}(s^\prime) = \sum_{i=1}^m \phi_i \mathcal{N}({\mu}_i, {\sigma}_i^2),
\end{flalign}
where $\phi_i: \mcs \rightarrow [0, 1]$ and $\sum_{i=1}^m \phi_i = 1$, $\mathcal{N}$ denotes the Gaussian distribution and $\boldsymbol{\mu} = (\mu_1, \cdots, \mu_m): \mcs \times \mca \rightarrow \mathbb{R}^m, \boldsymbol{\sigma} = (\sigma_1, \cdots, \sigma_m): \mcs \times \mca \rightarrow \mathbb{R}^m$ are the parameters. In this case, let $\boldsymbol{\xi} = (\boldsymbol{\mu}, \boldsymbol{\sigma})$.


We then evaluate the advantage function $A_{r, p^{r}_k}^{\pi_k}$ and the visitation distribution $d^{\pi_k}_{p^{r}_k}$ by performing policy $\pi_k$ under the transition kernel $p_k^r$. The intermediate policy $\pi_{k+\frac{1}{2}}$ is updated by solving the following optimization problem:
\begin{flalign}\label{eq:improve}
\max_{\pi\in \Pi_{\boldsymbol{\theta}}} \quad& \mathbb{E}_{\substack{s\sim d^{\pi_k}_{p^{r}_k}\\\hspace{-0.1cm}a\sim\pi}}[A_{r, p^{r}_k}^{\pi_k}(s, a)],\nn\\\text{s.t.}\  &\mathbb{E}_{s\sim d^{\pi_k}_{p^{r}_k}}[D_{KL}(\pi||\pi_k)(s)] \leq \delta.
\end{flalign}
Note that in \eqref{eq:improve}, $A_{r, p^{r}_k}^{\pi_k}$ and $d^{\pi_k}_{p^{r}_k}$ are estimated using the sample trajectories from the current policy $\pi_k$ under the transition kernel $p^{r}_k$, which can be easily obtained. We optimize the advantage function over a neighborhood of the current policy $\pi_k$. Therefore, the advantage function and visitation distribution under policy $\pi_k$ are good local approximations for all policies in this neighborhood.
For the tabular case, in the policy improvement step, we only need to find a policy $\pi$ that maximizes the expected value of $A_{r, p^{r}_k}^{\pi_k}(s, a)$ with $a\sim\pi$, which is linear in $\pi$, and satisfies the constraint on the expected $D_{KL}(\pi||\pi_k)(s)$ under the distribution $d^{\pi_k}_{p^{r}_k}$, which is convex in $\pi$. Therefore, \eqref{eq:improve} is a convex optimization problem and can be solved efficiently.

\textbf{Step 2: Projection.} By solving \eqref{eq:improve}, we obtain a policy $\pi_{k+\frac{1}{2}}$ that maximizes the advantage function in the neighborhood of current policy $\pi_k$. However, $\pi_{k+\frac{1}{2}}$ does not necessarily satisfy the constraint. In the projection step, we project the policy $\pi_{k+\frac{1}{2}}$ to the constraint set to obtain a constraint-satisfying policy $\pi_{k+1}$. We first estimate the worst-case transition kernel $p_k^{c}$ for the utility value function under the current policy $\pi_k$ using the projected gradient descent method:
\begin{flalign}
    p_k^{c, t+1} = \text{Proj}_{\cp}\big(p_k^{c, t} - \beta_t\nabla_{p}V_{c, p_k^{c, t}}^{\pi_k}(\rho)\big).
\end{flalign}
For the tabular case, $p_k^c$  can be obtained such that
\begin{flalign}\label{eq:worst_c}
V_{c, p_k^{c}}^{\pi_k}(\rho) = \min_{p\in\cp}V_{c, p}^{\pi_k}(\rho).
\end{flalign}
For the large/continuous state space, we parameterize the transition kernel as introduced in {Step 1}.

We then estimate $A_{c, p^{c}_k}^{\pi_k}, d^{\pi_k}_{p_k^{c}}$ using sample trajectories from $p_k^{c}$.
The projection step is achieved by solving
\begin{flalign}\label{eq:project}
\min_{\pi\in\Pi_{\boldsymbol{\theta}}} &\quad \mathbb{E}_{s\sim d^{\pi_k}_{p^{r}_k}}[D_{KL}(\pi||\pi_{k+\frac{1}{2}})(s)]\nn\\
\text{s.t.}\ & V_{c, p^{c}_k}^{\pi_k}(\rho) + \mathbb{E}_{\substack{s\sim d^{\pi_k}_{p_k^{c}}\\\hspace{-0.1cm} a\sim \pi}}[A_{c, p^{c}_k}^{\pi_k}(s, a)] \geq d.
\end{flalign}
In \eqref{eq:project}, the constraint $V_{c, p^{c}_k}^{\pi_k}(\rho) + \mathbb{E}_{s\sim d^{\pi_k}_{p_k^{c}}, a\sim \pi}[A_{c, p^{c}_k}^{\pi_k}(s, a)]$ is a local approximation for $V_c^\pi(\rho)$. For the tabular case, problem in \eqref{eq:project} is a convex optimization since the advantage function and the visitation distribution are obtained from the current policy $\pi_k$, and therefore, can be solved efficiently.

Unlike solving \eqref{eq:rcpo}, which might be infeasible when the current policy $\pi_k$ doesn't satisfy the constraint, our update rule consists of two convex optimization problems \eqref{eq:improve} and \eqref{eq:project},  the feasible set of which are much larger than \eqref{eq:rcpo}. 

\begin{algorithm}[tb]
   \caption{Robust Constrained Policy Optimization}
   \label{alg:rcpo}
\begin{algorithmic}
   \STATE {\bfseries Input:} step size $\delta,\{\beta_t\}_{t\geq 0}$, iteration time $K, T$, initial policy $\pi_0$
   \FOR{$k = 0, 1, \cdots, K-1$}
   \STATE Initialize $p_k^{r, 0}, p_k^{c, 0}$
   \FOR{$t = 0, 1, \cdots, T-1$}
   \STATE $p_k^{r, t+1} \leftarrow \text{Proj}_{\cp}\big(p_k^{r, t} - \beta_t\nabla_{p}V_{r, p_k^{r, t}}^{\pi_k}(\rho)\big)$
   \STATE $p_k^{c, t+1} \leftarrow \text{Proj}_{\cp}\big(p_k^{c, t} - \beta_t\nabla_{p}V_{c, p_k^{c, t}}^{\pi_k}(\rho)\big)$
   \ENDFOR 
   \STATE $p_k^{r} \leftarrow p_k^{r, T}$, $p_k^{c} \leftarrow p_k^{c, T}$
   \STATE Compute $A^{\pi_k}_{r, p^{r}_k}$, $A^{\pi_k}_{c, p^{c}_k}$, $d^{\pi_k}_{p^{r}_k}$, $d^{\pi_k}_{p^{c}_k}$
   \STATE Update $\pi_{k+\frac{1}{2}}$ according to \eqref{eq:improve}
   \STATE Update $\pi_{k+1}$ according to \eqref{eq:project}
   \ENDFOR
   \STATE {\bfseries Output:} $\pi_K$
\end{algorithmic}
\end{algorithm}


\subsection{Theoretical Results}
We first make the following assumption on the worst-case transition kernel.
\begin{assumption}\label{assumption:kernel}
We are able to find transition kernels $p_k^{r, \boldsymbol{\xi}}, p_k^{c, \boldsymbol{\xi}}$ such that 
\begin{flalign}\label{eq:kernel}
&\big|V_{r, p_k^{r, \boldsymbol{\xi}}}^{\pi_k}(\rho) - V_{r, p_k^r}^{\pi_k}(\rho)\big|\leq \epsilon,\nn\\&
\big|V_{c, p_k^{c, \boldsymbol{\xi}}}^{\pi_k}(\rho) - V_{c, p_k^c}^{\pi_k}(\rho)\big|\leq \epsilon.
\end{flalign}
\end{assumption}
This assumption can be satisfied under the tabular case using a direct parameterization of the transition kernel or under the case with a continuous state space if a large enough neural network is used to parameterize the transition kernel.

In the following theorem, we provide a lower bound on the worst-case reward improvement and an upper bound on the worst-case constraint violation for each iteration of our RCPO algorithm in \Cref{alg:rcpo}.
\begin{theorem}\label{theorem:feasible}
    Let $\epsilon_{r, p_{k+1}^{r}}^{\pi_{k+1}} = \max_s |\mathbb{E}_{a\sim\pi_{k+1}} A^{\pi_k}_{r, p_{k+1}^{r}}(s, a)|$ and $\epsilon_{c, p_{k+1}^{c}}^{\pi_{k+1}} = \max_s |\mathbb{E}_{a\sim\pi_{k+1}} A^{\pi_k}_{c, p_{k+1}^{c}}(s, a)|$. Under \Cref{assumption:kernel}, when the current policy $\pi_k$ satisfies the constraint in \eqref{eq:problem}, we have:
    \begin{flalign}
   & \text{Worst-case reward improvement: }\nn\\
    &V_r^{\pi_{k+1}}(\rho) - V_r^{\pi_k}(\rho) \geq - \frac{1}{1-\gamma}M\Big(2L_\pi + \frac{2\gamma \epsilon_{r, p_{k+1}^{r}}^{\pi_{k+1}}}{1-\gamma}\Big)\sqrt{\frac{\delta}{2}};\nn\\
    &\text{Constraint violation: }\nn\\
    &
    V_c^{\pi_{k+1}}(\rho) \geq d - \epsilon - \frac{1}{1-\gamma}M\Big(3L_\pi + \frac{2\gamma \epsilon_{c, p_{k+1}^{c}}^{\pi_{k+1}}}{1-\gamma}\Big)\sqrt{\frac{\delta}{2}},\nn
    \end{flalign}
    where $M = \sup_{p, p^\prime \in\cp}\|d^{\pi_k}_p/d^{\pi_k}_{p^\prime}\|_\infty$ is finite whenever $supp(\rho) = \mcs$ and $L_\pi = \frac{\sqrt{|\mca|}}{(1-\gamma)^2}$.
\end{theorem}
Theorem \ref{theorem:feasible} shows that we could adjust $\delta$ towards improved robust reward value function and smaller constraint violation. Moreover, for the large/continuous state space, our RCPO only incurs an additional degradation $\epsilon$ on constraint violation due to the worst-case transition kernel mismatch. For the tabular case, $\epsilon$ can be arbitrarily close to zero.
On the other hand, throughout the training, the policy $\pi_k$ may violate the constraint due to the random initialization or estimation errors. Therefore, in the following, we also characterize the performance of our algorithm when the current policy $\pi_k$ violates the constraint.

Let $b =d- V_c^{\pi_k}(\rho)$. The following theorem provides a lower bound on the worst-case reward improvement and an upper bound on the worst-case constraint violation.
\begin{theorem}\label{theorem:non-feasible}
    Under \Cref{assumption:kernel}, when the current policy $\pi_k$ violates the constraint in \eqref{eq:project}, we have:\\
Worst-case reward improvement: 
    \begin{flalign}
    &V_r^{\pi_{k+1}}(\rho) - V_r^{\pi_k}(\rho)\nn\\ & \geq - \frac{1}{1-\gamma}M\Big(2L_\pi + \frac{2\gamma \epsilon_{r, p_{k+1}^{r}}^{\pi_{k+1}}}{1-\gamma}\Big)\sqrt{\frac{\delta + b^2\alpha_{KL}}{2}};
    \end{flalign}
    Constraint violation: 
    \begin{flalign}
    &V_c^{\pi_{k+1}}(\rho) \geq d - \epsilon - \frac{1}{1-\gamma}M\Big(3L_\pi + \frac{2\gamma \epsilon_{c, p_{k+1}^{c}}^{\pi_{k+1}}}{1-\gamma}\Big)\nn\\
    & \times\sqrt{\frac{\delta + b^2\alpha_{KL} + bM^\prime\sqrt{\frac{\alpha_{KL}}{2}}}{2}},
    \end{flalign}
    where $\alpha_{KL} = \frac{1}{2\boldsymbol{h}^\top\boldsymbol{H}^{-1}\boldsymbol{h}}$, $\boldsymbol{h}$ and $\boldsymbol{H}$ are defined  in \eqref{eq:grad} and \eqref{eq:hess}, $M^\prime < \infty$ is some constant.
\end{theorem}
Theorem \ref{theorem:non-feasible} characterizes the performance of our algorithm when the current policy $\pi_k$ is infeasible. A small $b$, i.e., the current policy $\pi_k$ only violates the constraint slightly, leads to a better robust reward improvement and a smaller constraint violation. 
If the current policy $\pi_k$ satisfies the constraint, i.e., $b = 0$, Theorem \ref{theorem:non-feasible} reduces to Theorem \ref{theorem:feasible}.  The misspecified worst-case transition kernel only incurs an additional performance degradation $\epsilon$ on the constraint violation for large/continuous state space. For the tabular case, $\epsilon$ can be arbitrarily close to zero.

\section{Practical Implementation}\label{sec:practical}

In this section, we provide a practical implementation of \Cref{alg:rcpo} to tackle the computational challenge. 

To update the policy efficiently, for a small step size $\delta$, we approximate the objective functions and constraints in the optimization problems \eqref{eq:improve} and \eqref{eq:project} using their Taylor expansions. 
Let
\begin{flalign}\label{eq:grad}
 \boldsymbol{g} &= \nabla_{\boldsymbol{\theta}}\mathbb{E}_{s\sim d^{\pi_k}_{p^{r}_{k, \boldsymbol{\xi}}}, a\sim \pi}[A_{r, p^{r}_{k, \boldsymbol{\xi}}}^{\pi_k}(s, a)],\nn\\
 \boldsymbol{h} &= \nabla_{\boldsymbol{\theta}}\mathbb{E}_{s\sim d^{\pi_k}_{p^{c}_{k, \boldsymbol{\xi}}}, a\sim \pi}[A_{c, p^{c}_{k, \boldsymbol{\xi}}}^{\pi_k}(s, a)]
\end{flalign}
be the gradient of the reward advantage function and the gradient of the utility advantage function, respectively.
Moreover, let
\begin{flalign}\label{eq:hess}
\boldsymbol{H} = \nabla^2_{\boldsymbol{\theta}}{ \mathbb{E}_{s\sim d^{\pi_k}_{p^{r}_{k, \boldsymbol{\xi}}}}\big[D_{KL}(\pi||\pi_k)(s)\big]}
\end{flalign}
be the Hessian matrix of the KL divergence. We then develop the following practical implementation for our RCPO algorithm.

\textbf{Step 1: Robust Policy Improvement.} We first estimate the worst-case transition kernel $p_k^{r}$ for the current policy $\pi_k$. 
We iteratively update the parameterized transition kernel $p_{k, \boldsymbol{\xi}}^{r, t}$ using the following projected gradient descent method:
\begin{flalign}\label{eq:tract_grad}
    p_{k, \boldsymbol{\xi}}^{r, t+1} = \text{Proj}_{\cp}\left(p_{k, \boldsymbol{\xi}}^{r, t} - \beta_t\nabla_{p}V_{r, p_{k, \boldsymbol{\xi}}^{r, t}}^{\pi_k}(\rho)\right).
\end{flalign}

We then use the first-order approximation for the objective function and the second-order approximation for the KL divergence constraint at the current policy $\pi_k$ in \eqref{eq:improve}. Let $\boldsymbol{\theta}_k$ denote the parameter of policy $\pi_k$. The parameter of the intermediate policy $\pi_{k+\frac{1}{2}}$ is updated by solving the following practical formulation for \eqref{eq:improve}:
\begin{flalign}\label{eq:tract_imporve}
\max_{\boldsymbol{\theta}} \boldsymbol{g}^\top (\boldsymbol{\theta} - \boldsymbol{\theta}_k),\ \text{s.t.}\ \frac{1}{2}(\boldsymbol{\theta}-\boldsymbol{\theta}_k)^\top \boldsymbol{H}(\boldsymbol{\theta}-\boldsymbol{\theta}_k)\leq \delta.
\end{flalign}
The objective function of \eqref{eq:tract_imporve} is linear in $\boldsymbol{\theta}$ and the constraint is quadratic in $\boldsymbol{\theta}$. Therefore, problem \eqref{eq:tract_imporve} can be easily solved.

\textbf{Step 2: Projection.} For the projection step, we first estimate the worst-case transition kernel $p_k^{c}$ for utility. Similarly, we iteratively update the parameterized transition kernel $p_{k, \boldsymbol{\xi}}^{c, t}$ using the following projected gradient descent method:
\begin{flalign}
   p_{k, \boldsymbol{\xi}}^{c, t+1} = \text{Proj}_{\cp}\big(p_{k, \boldsymbol{\xi}}^{c, t} - \beta_t\nabla_{p}V_{c, p_{k, \boldsymbol{\xi}}^{c, t}}^{\pi_k}(\rho)\big).
\end{flalign}

We then approximate the objective function in \eqref{eq:project} by its second order expansion and approximate the constraint in \eqref{eq:project} by its first order expansion. The parameter of the policy $\pi_{k+1}$ is then updated by solving the following problem:
\begin{flalign}\label{eq:tract_proj}
\min_{\boldsymbol{\theta}} &\frac{1}{2} (\boldsymbol{\theta} - \boldsymbol{\theta_{k+\frac{1}{2}}})^\top \boldsymbol{H}(\boldsymbol{\theta} - \boldsymbol{\theta_{k+\frac{1}{2}}}) \nn\\ \text{s.t.}\ &\boldsymbol{h}^\top(\boldsymbol{\theta} - \boldsymbol{\theta}_k) + b \leq 0.
\end{flalign}

The problems in \eqref{eq:tract_imporve} and \eqref{eq:tract_proj} can be solved by convex programming \cite{yang2019projection}. We have the following update rule for each policy update.
\begin{flalign}\label{eq:update_approx}
\boldsymbol{\theta}_{k+1} &= \boldsymbol{\theta}_k + \sqrt{\frac{2\delta}{\boldsymbol{g}^\top \boldsymbol{H}^{-1} \boldsymbol{g}}} \boldsymbol{H}^{-1} \boldsymbol{g}\nn\\& - \max\Bigg(\frac{\sqrt{\frac{2\delta}{\boldsymbol{g}^\top \boldsymbol{H}^{-1} \boldsymbol{g}}}\boldsymbol{h} \boldsymbol{H}^{-1} \boldsymbol{g} + b}{\boldsymbol{h}^\top \boldsymbol{H}^{-1} \boldsymbol{h}}, 0\Bigg) \boldsymbol{H}^{-1} \boldsymbol{h}.
\end{flalign}

In this way, our RCPO algorithm can be implemented efficiently using the above described approximation for large-scale problems.

\section{Experiments}
To validate the proposed algorithm, we compare it with several baseline algorithms (PCPO \cite{yang2020projection}, RVI \cite{iyengar2005robust}, and CPO \cite{achiam2017constrained}) in the setting of tabular and deep cases, while using different environments such as the gambler problem \cite{sutton2018reinforcement, zhou2021finite, shi2022distributionally}, the $N$-chain problem \cite{wang2022robust}, the Frozen-Lake problem \cite{brockman2016openai} and the Point Gather in Mujoco \cite{achiam2017constrained, yang2020projection}. 

\subsection{Tabular Case}
In the setting of tabular cases, we evaluate the performance of our algorithm in the environment of gambler problem, N-chain problem and Frozen-Lake problem, where both state and action spaces are finite. We compare our algorithm with two baselines: the PCPO \cite{yang2020projection} and the model-based robust value iteration (RVI) \cite{iyengar2005robust}. The PCPO learns an optimal policy subjecting to the constraints under the nominal transition kernels without considering the model mismatch. The model-based RVI directly optimizes the unconstrained robust reward objective, which serves as an upper bound of the reward value function for the robust constrained problem. We consider the KL divergence uncertainty set. For each problem, we run the algorithms for $5$ independent times and plot the mean of the reward and utility along with their standard deviation as a function of the number of iterations. The detailed environments descriptions can be found in Appendix \ref{sec:appendix_sim}.

\begin{figure}[!htb]
\vskip -0.2in
\begin{center}
\subfigure{
\label{fig:gambler_r}
\includegraphics[width=0.47\linewidth]{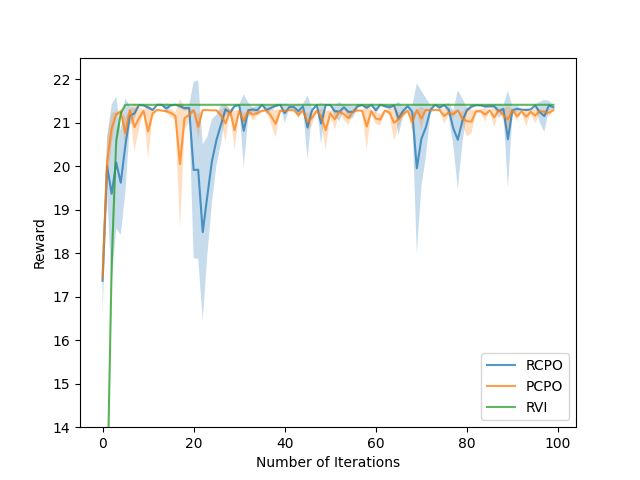}}
\subfigure{\label{fig:gambler_c}
\includegraphics[width=0.47\linewidth]{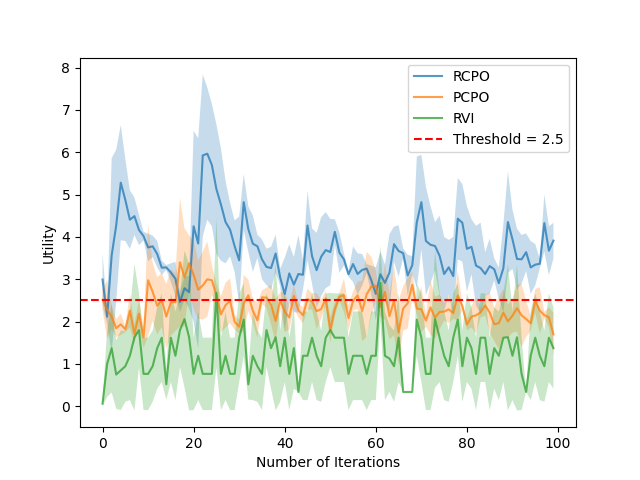}}
\caption{Gambler Problem}
\end{center}
\vskip -0.05in
\end{figure}

For the gambler problem, the threshold for the constraint is $2.5$. It can be seen from Fig. \ref{fig:gambler_c} that our RCPO always satisfies the constraint during the training while both the PCPO and RVI violate the constraints. Moreover, the reward of our RCPO in Fig. \ref{fig:gambler_r} is close to the reward of RVI, which is the best achievable reward for the unconstrained robust RL. Therefore, our algorithm learns a policy that satisfies the worst-case constraint on the utility and achieves optimal reward objective.

\begin{figure}[!htb]
\vskip -0.2in
\begin{center}
\subfigure{
\label{fig:nchain_r}
\includegraphics[width=0.47\linewidth]{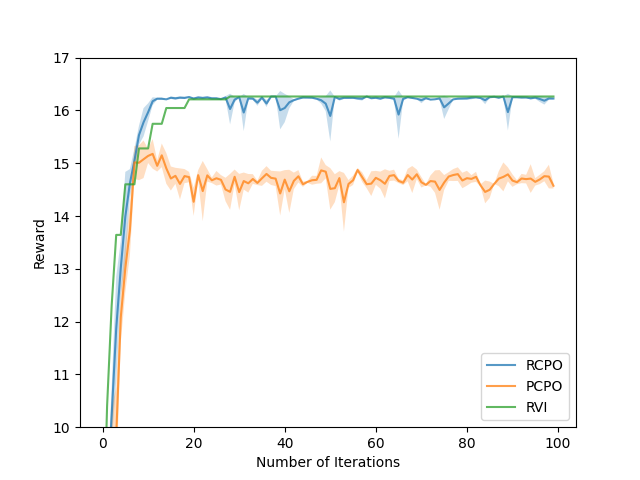}}
\subfigure{\label{fig:nchain_c}
\includegraphics[width=0.47\linewidth]{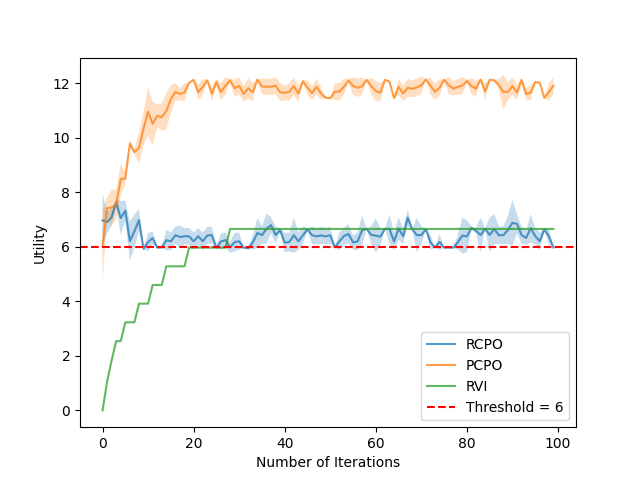}}
\caption{$N$-chain Problem}
\end{center}
\end{figure}

For the $N$-chain problem, the threshold is set to be $6$. From Fig. \ref{fig:nchain_c}, it can be seen that all three algorithms satisfy the constraint, indicating that the constraint is easy to satisfy for this problem. However, in Fig. \ref{fig:nchain_r}, PCPO converges to a sub-optimal policy while our RCPO has a similar convergence rate as the unconstrained RVI.

\begin{figure}[!htb]
\vskip -0.1in
\begin{center}
\subfigure{
\label{fig:frozen_r}
\includegraphics[width=0.47\linewidth]{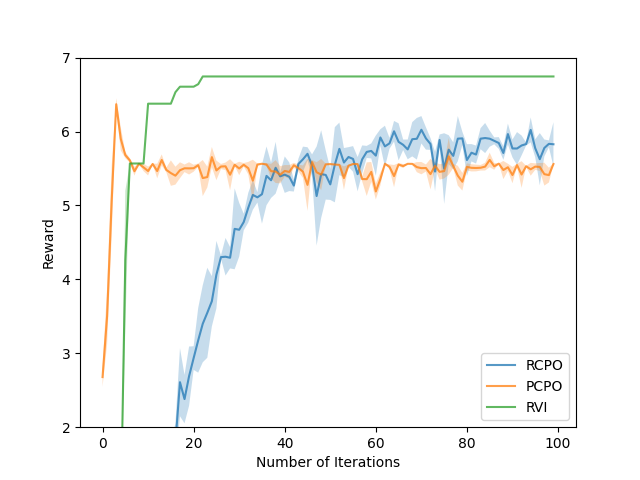}}
\subfigure{\label{fig:frozen_c}
\includegraphics[width=0.47\linewidth]{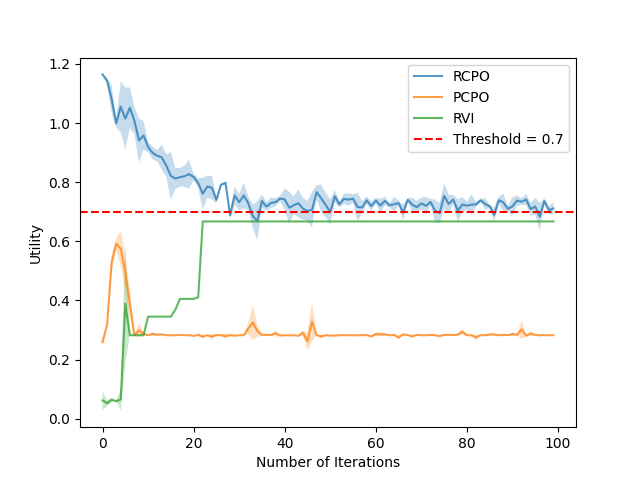}}
\caption{Frozen-Lake Problem}
\end{center}
\vskip -0.05in
\end{figure}

For the Frozen-Lake problem, the threshold is set to be $0.7$. From Fig. \ref{fig:frozen_c}, it can be seen that only RCPO satisfies the constraint. Both PCPO and RVI violate the constraint during training. Moreover, in Fig. \ref{fig:frozen_r}, it can be seen that our RCPO obtain more reward than PCPO, which demonstrates the effectiveness and robustness of our algorithm.

\subsection{Deep Case}
\begin{figure}[!htb]
\vskip -0.2in
\begin{center}
\subfigure{
\label{fig:point_r}
\includegraphics[width=0.47\linewidth]{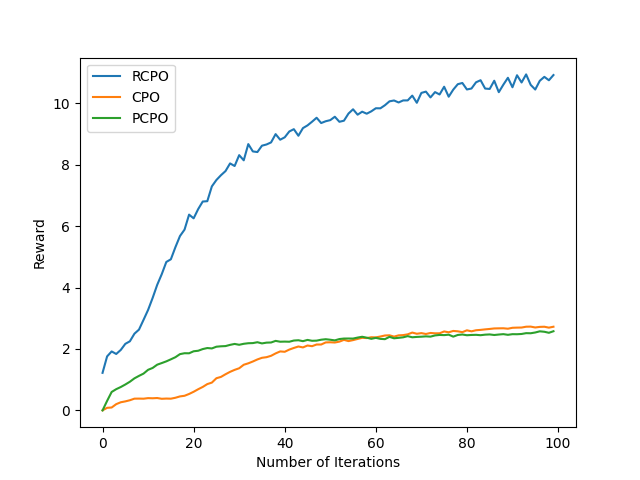}}
\subfigure{\label{fig:point_c}
\includegraphics[width=0.47\linewidth]{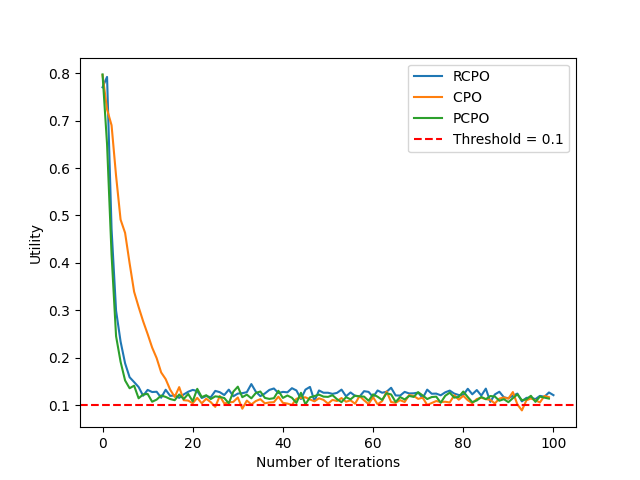}}
\caption{Point Gather}
\end{center}
\vskip -0.1in
\end{figure}

In the setting of the deep case, we incorporate our algorithm into deep neural networks for tackling high-dimensional spaces (e.g. continuous state spaces). We compare the proposed RCPO with CPO \cite{achiam2017constrained} and PCPO \cite{yang2020projection}. We use the same neural network policies with two hidden layers of size (64, 32) in all three algorithms. We adopt a Mujoco-based environment, Point Gather task with safety constraints \cite{achiam2017constrained}, which is a well-recognized constrained MDP environment. We use the following hyper-parameters for training RCPO: discounted factor $= 0.995$, learning step size $= 0.001$, batch size $= 50,000$, and utility-constrained threshold $= 0.1$. To provide fair comparisons, we use the same hyper-parameters for training baseline algorithms. The experiments are implemented in rllab \cite{duan2016benchmarking}, a tool for developing and evaluating RL algorithms. To introduce model uncertainties into the environment, we use Gaussian noise to perturb the environment and evaluate the performance of three algorithms under the perturbed environment. It can be seen from Fig. \ref{fig:point_c} that the rewards of RCPO are much higher than these two non-robust algorithms under model uncertainty, which demonstrates the robustness of our algorithm to model uncertainty when incorporating deep neural networks. Meanwhile, the well-trained RCPO policy satisfies the utility constraint. In summary, RCPO is able to provide efficient, robust, and constraint-satisfied policies in environments with continuous spaces by incorporating deep neural networks.

\section{Conclusion}
In this paper, we study the problem of constrained reinforcement learning under model mismatch. The goal is to maximize the worst-case reward over the uncertainty set subject to a constraint that the utility function for all transition kernels in the uncertainty set shall be above a prescribed threshold. We propose a robust constrained policy optimization (RCPO) algorithm, which consists of several novel technical developments than the CPO algorithm \cite{achiam2017constrained} for the non-robust standard CMDP problem. One result that may of independent interest is a robust performance difference lemma that bound the different between the robust value functions of two policies. Our algorithm is applicable to large scale MDPs, and has theoretical guarantees on worst-case reward improvement and constraint violation at each iteration during the training.  We further provide an efficient approximation for the purpose of practical implementation of our algorithm. Numerical experiments on demonstrate the effectiveness and robustness of our algorithm under model mismatch.

\newpage
\section{Impact Statements}
This paper presents work whose goal is to advance the field of Reinforcement Learning. There are many potential societal consequences of our work, none which we feel must be specifically highlighted here.
\bibliography{ref}
\bibliographystyle{icml2024}

\newpage
\appendix
\onecolumn

\section{Review of Constrained Policy Optimization}\label{app:cpo}
In this section, we provide an overview of the CPO method developed in \cite{achiam2017constrained}. 
Recall that in the standard (non-robust) RL setting, the value function difference between two policies $\pi, \pi^\prime$ can be written as \cite{kakade2002approximately}:
\begin{flalign}\label{eq:per_diff}
V_{r, p}^{\pi\prime}(\rho) - V_{r, p}^{\pi}(\rho) = \frac{1}{1-\gamma}\mathbb{E}_{\substack{s\sim d^{\pi^\prime}_p \\ a\sim \pi^\prime}} [A^\pi_{r, p}(s, a)],
\end{flalign}
where $A^\pi_{r, p}(s, a) = Q^\pi_{r, p}(s, a) - V^\pi_{r, p}(s)$ is the reward advantage function. 
In \cite{achiam2017constrained}, the above result was further extended to the following one:
\begin{flalign}\label{eq:bound}
&V_{r, p}^{\pi^\prime}(\rho) - V_{r, p}^{\pi}(\rho)\\
& \geq \frac{1}{1-\gamma} \mathbb{E}_{\substack{s\sim d^{\pi}_p\\a\sim \pi^\prime}}\left[A^\pi_{r, p}(s, a)-\frac{2\gamma\epsilon^{\pi^\prime}_r}{1-\gamma}\sqrt{\frac{1}{2}D_{KL}(\pi^\prime\|\pi)(s)}\right],\nn
\end{flalign}
where $\epsilon^{\pi^\prime}_r=\max_s |\mathbb{E}_{a\sim\pi^\prime}[A^\pi_{r, p}(s, a)]|$.

\Cref{eq:bound} connects the performance difference between two policies to an average divergence between them. Compared with \eqref{eq:per_diff}, the expectation is taken with respect to $d^{\pi}_p$ in \eqref{eq:bound} instead of $d^{\pi^\prime}_p$. When $\pi^\prime$ is close to $\pi$, $D_{KL}(\pi^\prime\|\pi)(s)$ is small and $d^{\pi^\prime}_p$ is close to $d^{\pi}_p$. Therefore, the right hand side of \eqref{eq:bound} is a good local approximation for the performance difference $V_{r, p}^{\pi^\prime} - V_{r, p}^{\pi}$. The trust region method for unconstrained RL was proposed \cite{schulman2015trust, schulman2015high} based on this approximation and  provides monotonic improvement for the reward value function.

For the utility value function, we have the following equivalent expression:
\begin{flalign}\label{eq:utility_bound}
  &V_{c, p}^{\pi^\prime}(\rho) - V_{c, p}^{\pi}(\rho)\\
  & \geq \frac{1}{1-\gamma} \mathbb{E}_{\substack{s\sim d^{\pi}_p\\a\sim \pi^\prime}}\left[A^\pi_{c, p}(s, a)-\frac{2\gamma\epsilon^{\pi^\prime}_c}{1-\gamma}\sqrt{\frac{1}{2}D_{KL}(\pi^\prime\|\pi)(s)}\right],  \nn
\end{flalign}
where $\epsilon^{\pi^\prime}_c=\max_s |\mathbb{E}_{a\sim\pi^\prime}[A^\pi_{c, p}(s, a)]|$ and $A^\pi_{c, p}(s, a) = Q^\pi_{c, p}(s, a) - V^\pi_{c, p}(s)$ is the utility advantage function. The right hand side of \eqref{eq:utility_bound} can be used as an approximation for $V_{c, p}^{\pi^\prime}(\rho) - V_{c, p}^{\pi}(\rho)$. By applying the trust region methods to CMDPs, the constrained policy optimization (CPO) was proposed in \cite{achiam2017constrained}, where the policy is updated by solving the following optimization problem.
\begin{flalign}\label{eq:cpo}
    \pi_{k+1}&= \argmax_{\pi}\mathbb{E}_{\substack{s\sim d^{\pi_k}_p\\a\sim \pi}}\Big[A^{\pi_k}_{r, p}(s, a)\Big]\nn\\
    \text{s.t.}&\ V_{c, p}^{\pi_k}(\rho) +\frac{1}{1-\gamma} \mathbb{E}_{\substack{s\sim d^{\pi_k}_p\\a\sim \pi}}\Big[A^{\pi_k}_{c, p}(s, a)\Big] \geq d, \nn\\& \mathbb{E}_{s\sim d^{\pi_k}_p}[D_{KL}(\pi\|\pi_k)(s)]\leq \delta.
\end{flalign}
When the current policy $\pi_k$ satisfies the constraint, this update rule leads to a policy that has performance improvement and approximate satisfaction of constraints \cite{achiam2017constrained}. Note that the expectation in the optimization problem \eqref{eq:cpo} is taken with respect to $d^{\pi_k}_p$. The optimization problem \eqref{eq:cpo} depends on $\pi$ only through the distribution of the current action $a$, which can thus be optimized efficiently.
\section{Proof of Lemma \ref{lemma:rob_perf}}
For any policies $\pi, \pi^\prime$, we have that
\begin{flalign}
    V_r^{\pi^\prime}(\rho) - V_r^\pi(\rho)
    &= V_{r, p_{\pi^\prime}^{r}}^{\pi^\prime}(\rho) - V_{r, p_{\pi}^{r}}^{\pi}(\rho)\nn\\& \geq V_{r, p_{\pi^\prime}^{r}}^{\pi^\prime}(\rho) - V_{r, p_{\pi^\prime}^{r}}^{\pi}(\rho)\nn\\&\geq
    \frac{1}{1-\gamma} \mathbb{E}_{\substack{s\sim d^{\pi}_{p_{\pi^\prime}^{r}}\\a\sim \pi^\prime}}\Big[A^\pi_{r, {p_{\pi^\prime}^{r}}}(s, a) -\frac{2\gamma\epsilon^{\pi^\prime}_{r, p_{\pi^\prime}^{r}}}{1-\gamma}\sqrt{\frac{1}{2}D_{KL}(\pi^\prime\|\pi)(s)}\Big],
\end{flalign} 
where the first inequality is because $V_{r,p_{\pi}^{r}}^\pi(\rho) = \min_{p\in\cp}V_{r,p}^\pi(\rho)$ and the second inequality follows from Theorem 1 in \cite{achiam2017constrained}.

\section{Proof of Theorem \ref{theorem:feasible}}
We first prove the follow Lemma.
\begin{lemma}\label{lemma:kl}
If the current policy $\pi_k$ satisfies the constraint and the constraint set is closed and convex under the policy parameterization, then under the KL divergence projection, we have that 
\begin{flalign}\label{eq:large_kl}
\mathbb{E}_{s\sim d^{\pi_k}_{p^{r, \boldsymbol{\xi}}_k}}[D_{KL}(\pi_{k+1}||\pi_k)(s)] \leq \delta.
\end{flalign}
\end{lemma}
\begin{proof}
Note that the constraint in \eqref{eq:project} is linear in $\pi$. Therefore, the constraint set is closed and convex. Since $\pi_k$ lies in the constraint set and $\pi_{k+1}$ is the projection of $\pi_{k+\frac{1}{2}}$ onto the constraint set, from the Bregmann divergence projection inequality, we have that 
\begin{flalign}
\mathbb{E}_{s\sim d^{\pi_k}_{p^{r, \boldsymbol{\xi}}_k}}[D_{KL}(\pi_{k}||\pi_{k+\frac{1}{2}})(s)] \geq \mathbb{E}_{s\sim d^{\pi_k}_{p^{r, \boldsymbol{\xi}}_k}}[D_{KL}(\pi_{k}||\pi_{k+1})(s)] + \mathbb{E}_{s\sim d^{\pi_k}_{p^{r, \boldsymbol{\xi}}_k}}[D_{KL}(\pi_{k+1}||\pi_{k+\frac{1}{2}})(s)]. 
\end{flalign}
Since the KL divergence is non-negative, we have that 
\begin{flalign}
\mathbb{E}_{s\sim d^{\pi_k}_{p^{r, \boldsymbol{\xi}}_k}}[D_{KL}(\pi_{k}||\pi_{k+\frac{1}{2}})(s)] \geq \mathbb{E}_{s\sim d^{\pi_k}_{p^{r, \boldsymbol{\xi}}_k}}[D_{KL}(\pi_{k}||\pi_{k+1})(s)].
\end{flalign}
When $\delta$ is small, the KL divergence is asymptotically symmetric. Therefore, we have that
\begin{flalign}
\mathbb{E}_{s\sim d^{\pi_k}_{p^{r, \boldsymbol{\xi}}_k}}[D_{KL}(\pi_{k+1}\|\pi_{k})(s)] \leq \mathbb{E}_{s\sim d^{\pi_k}_{p^{r, \boldsymbol{\xi}}_k}}[D_{KL}(\pi_{k+\frac{1}{2}}\|\pi_{k})(s)] \leq \delta.
\end{flalign}
\end{proof}

With Lemma \ref{lemma:kl}, we are ready to prove Theorem \ref{theorem:feasible}.

\begin{proof}
From Lemma \ref{lemma:rob_perf}, we have that for the reward improvement,
\begin{flalign}
V_r^{\pi_{k+1}}(\rho) - V_r^{\pi_k}(\rho)&
\geq \frac{1}{1-\gamma} \mathbb{E}_{\substack{s\sim d^{\pi_k}_{p_{{k+1}}^{r}}\\a\sim \pi_{k+1}}}\Big[A^{\pi_k}_{r, {p_{{k+1}}^{r}}}(s, a)-\frac{2\gamma\epsilon^{\pi_{k+1}}_{r,p_{k+1}^{r}}}{1-\gamma}\sqrt{\frac{1}{2}D_{KL}(\pi_{k+1}\|\pi_k)(s)}\Big].
\end{flalign}
Note that $A^{\pi_k}_{r, {p_{{k+1}}^{r}}}(s, a) = Q^{\pi_k}_{r, {p_{{k+1}}^{r}}}(s, a) - V^{\pi_k}_{r, {p_{{k+1}}^{r}}}(s)$ is Lipschitz in $\pi_k$ \cite{wang2023policy}. We have that there exists $L_\pi$ such that 
\begin{flalign}
|A^{\pi_k}_{r, {p_{{k+1}}^{r}}}(s, a) - A^{{k+1}}_{r, {p_{{k+1}}^{r}}}(s, a)| \leq L_\pi\|\pi_{k+1}(s) - \pi_k(s)\|_1.
\end{flalign}
We then have that
\begin{flalign}\label{eq:rew}
&\frac{1}{1-\gamma} \mathbb{E}_{\substack{s\sim d^{\pi_k}_{p_{{k+1}}^{r}}\\a\sim \pi_{k+1}}}\Big[A^{\pi_k}_{r, {p_{k+1}^{r}}}(s, a)-\frac{2\gamma\epsilon^{\pi_{k+1}}_{r,p_{k+1}^{r}}}{1-\gamma}\sqrt{\frac{1}{2}D_{KL}(\pi_{k+1}\|\pi_k)(s)}\Big]\nn\\&\geq \frac{1}{1-\gamma} \mathbb{E}_{\substack{s\sim d^{\pi_k}_{p_{{k+1}}^{r}}\\a\sim \pi_{k+1}}}\Big[A^{\pi_{k+1}}_{r, {p_{k+1}^{r}}}(s, a) - L_\pi\|\pi_{k+1}(s)-\pi_k(s)\|_1-\frac{2\gamma\epsilon^{\pi_{k+1}}_{r,p_{k+1}^{r}}}{1-\gamma}\sqrt{\frac{1}{2}D_{KL}(\pi_{k+1}\|\pi_k)(s)}\Big]\nn\\& = \frac{1}{1-\gamma} \mathbb{E}_{s\sim d^{\pi_k}_{p_{k+1}^{r}}}\Big[ - L_\pi\|\pi_{k+1}(s)-\pi_k(s)\|_1-\frac{2\gamma\epsilon^{\pi_{k+1}}_{r,p_{k+1}^{r}}}{1-\gamma}\sqrt{\frac{1}{2}D_{KL}(\pi_{k+1}\|\pi_k)(s)}\Big],
\end{flalign}
where the equality is due to the fact that $\mathbb{E}_{a\sim \pi_{k+1}}\big[A^{\pi_{k+1}}_{r, {p_{k+1}^{r}}}(s, a)\big] = \mathbb{E}_{a\sim \pi_{k+1}}\big[Q^{\pi_{k+1}}_{r, {p_{k+1}^{r}}}(s, a) - V^{\pi_{k+1}}_{r, {p_{k+1}^{r}}}(s) \big] = 0$.
Since $\|\pi_{k+1}(s)-\pi_k(s)\|_1  = 2D_{TV}(\pi_{k+1}\|\pi_k)(s)\leq \sqrt{2D_{KL}(\pi_{k+1}\|\pi_k)(s)}$ \cite{csiszar2011information}, we have that 
\begin{flalign}\label{eq:app_reward}
&\frac{1}{1-\gamma} \mathbb{E}_{s\sim d^{\pi_k}_{p_{k+1}^{r}}}\Big[ - L_\pi\|\pi_{k+1}(s)-\pi_k(s)\|_1-\frac{2\gamma\epsilon^{\pi_{k+1}}_{r,p_{k+1}^{r}}}{1-\gamma}\sqrt{\frac{1}{2}D_{KL}(\pi_{k+1}\|\pi_k)(s)}\Big]\nn\\& \geq \frac{1}{1-\gamma} \mathbb{E}_{s\sim d^{\pi_k}_{p_{k+1}^{r}}}\Big[-\big(2L_\pi + \frac{2\gamma\epsilon^{\pi_{k+1}}_{r,p_{k+1}^{r}}}{1-\gamma}\big)\sqrt{\frac{1}{2}D_{KL}(\pi_{k+1}\|\pi_k)(s)}\Big]\nn\\&\overset{(a)}{\geq} \frac{1}{1-\gamma} \mathbb{E}_{s\sim d^{\pi_k}_{p^{r, \boldsymbol{\xi}}_k}}\Big[-M\big(2L_\pi + \frac{2\gamma\epsilon^{\pi_{k+1}}_{r,p_{k+1}^{r}}}{1-\gamma}\big)\sqrt{\frac{1}{2}D_{KL}(\pi_{k+1}\|\pi_k)(s)}\Big]\nn\\&\overset{(b)}{\geq} -\frac{1}{1-\gamma} M \big(2L_\pi + \frac{2\gamma\epsilon^{\pi_{k+1}}_{r,p_{k+1}^{r}}}{1-\gamma}\big)\sqrt{\frac{\delta}{2}},
\end{flalign}
where $(a)$ is due to the fact that $M = \sup_{p, p^\prime \in\cp}\|d^{\pi_k}_p/d^{\pi_k}_{p^\prime}\|_\infty$ is finite and $(b)$ is from Lemma \ref{lemma:kl} and Jensen's inequality. 

To characterize the constraint violation, we first have that 
\begin{flalign}
V_{c, p^{c, \boldsymbol{\xi}}_k}^{\pi_k}(\rho) + \mathbb{E}_{\substack{s\sim d^{\pi_k}_{p_k^{c, \boldsymbol{\xi}}}\\ a\sim \pi_{k+1}}}[A_{c, p^{c, \boldsymbol{\xi}}_k}^{\pi_k}(s, a)] \geq d.
\end{flalign}
and 
\begin{flalign}
V_c^{\pi_{k+1}}(\rho) - V_c^{\pi_k}(\rho)&= V_{c, p_{k+1}^{c}}^{\pi_{k+1}}(\rho) - V_{c, p_{\pi_k}^{c}}^{\pi_k}(\rho)\nn\\& \geq V_{c, p_{k+1}^{c}}^{\pi_{k+1}}(\rho) - V_{c, p_{k+1}^{c}}^{\pi_k}(\rho)\nn\\&\geq \frac{1}{1-\gamma} \mathbb{E}_{\substack{s\sim d^{\pi_k}_{p_{k+1}^{c}}\\a\sim \pi_{k+1}}}\Big[A^{\pi_k}_{c, {p_{k+1}^{c}}}(s, a)-\frac{2\gamma\epsilon^{\pi_{k+1}}_{c,p_{k+1}^{c}}}{1-\gamma}\sqrt{\frac{1}{2}D_{KL}(\pi_{k+1}\|\pi_k)(s)}\Big].
\end{flalign}

Following the proof of \eqref{eq:app_reward}, we have that
\begin{flalign}
V_c^{\pi_{k+1}}(\rho) &\geq d - (V_{c, p^{c, \boldsymbol{\xi}}_k}^{\pi_k}(\rho) - V_c^{\pi_k}(\rho)) - \mathbb{E}_{\substack{s\sim d^{\pi_k}_{p_k^{c, \boldsymbol{\xi}}}\\ a\sim \pi_{k+1}}}[A_{c, p^{c, \boldsymbol{\xi}}_k}^{\pi_k}(s, a)] \nn\\&\quad + \frac{1}{1-\gamma} \mathbb{E}_{\substack{s\sim d^{\pi_k}_{p_{k+1}^{c}}\\a\sim \pi_{k+1}}}\Big[A^{\pi_k}_{c, {p_{k+1}^{c}}}(s, a)-\frac{2\gamma\epsilon^{\pi_{k+1}}_{c,p_{k+1}^{c}}}{1-\gamma}\sqrt{\frac{1}{2}D_{KL}(\pi_{k+1}\|\pi_k)(s)}\Big]\nn\\& \geq d - \epsilon - \mathbb{E}_{\substack{s\sim d^{\pi_k}_{p_k^{c, \boldsymbol{\xi}}}\\ a\sim \pi_{k+1}}}\big[A_{c, p^{c}_k}^{\pi_{k+1}}(s, a) + L_\pi\|\pi_{k+1}(s)-\pi_k(s)\|_1\big]\nn\\&\quad + \frac{1}{1-\gamma} \mathbb{E}_{\substack{s\sim d^{\pi_k}_{p_{k+1}^{c}}\\a\sim \pi_{k+1}}}\Big[A^{\pi_{k+1}}_{c, {p_{k+1}^{c}}}(s, a) - L_\pi\|\pi_{k+1}(s)-\pi_k(s)\|_1-\frac{2\gamma\epsilon^{\pi_{k+1}}_{c,p_{k+1}^{c}}}{1-\gamma}\sqrt{\frac{1}{2}D_{KL}(\pi_{k+1}\|\pi_k)(s)}\Big]\nn\\& = d - \epsilon - \mathbb{E}_{s\sim d^{\pi_k}_{p_k^{c, \boldsymbol{\xi}}}}[L_\pi\|\pi_{k+1}(s)-\pi_k(s)\|_1]\nn\\&\quad + \frac{1}{1-\gamma} \mathbb{E}_{s\sim d^{\pi_k}_{p_{k+1}^{c}}}\Big[ - L_\pi\|\pi_{k+1}(s)-\pi_k(s)\|_1-\frac{2\gamma\epsilon^{\pi_{k+1}}_{c,p_{k+1}^{c}}}{1-\gamma}\sqrt{\frac{1}{2}D_{KL}(\pi_{k+1}\|\pi_k)(s)}\Big]\nn\\& \geq d -\epsilon + \frac{1}{1-\gamma} \mathbb{E}_{s\sim d^{\pi_k}_{p_{k}^{c, \boldsymbol{\xi}}}}\Big[-M\big(3L_\pi + \frac{2\gamma\epsilon^{\pi_{k+1}}_{c,p_{k+1}^{c}}}{1-\gamma}\big)\sqrt{\frac{1}{2}D_{KL}(\pi_{k+1}\|\pi_k)(s)}\Big]\nn\\&\geq d - \epsilon - \frac{1}{1-\gamma} M \big(3L_\pi + \frac{2\gamma\epsilon^{\pi_{k+1}}_{c,p_{k+1}^{c}}}{1-\gamma}\big)\sqrt{\frac{\delta}{2}}.
\end{flalign}
This completes the proof.
\end{proof}

\section{Proof of Theorem \ref{theorem:non-feasible}}
We first provide an upper bound on the KL divergence between $\pi_k$ and $\pi_{k+1}$ in the following lemma. We then follow the proof of Theorem \ref{theorem:feasible} to prove Theorem \ref{theorem:non-feasible}.

\begin{lemma}\label{lemma:vio}
If the current policy $\pi_k$ violates the constraint and the constraint set is convex and closed under the policy parameterization, let $b = V_c^\pi(\rho) -d$, then under the KL divergence projection, we have that 
\begin{flalign}
\mathbb{E}_{s\sim d^{\pi_k}_{p^{r, \boldsymbol{\xi}}_k}}[D_{KL}(\pi_{k+1}||\pi_k)(s)] \leq \delta + b^2\alpha_{KL} + bM^\prime\sqrt{\frac{\alpha_{KL}}{2}},
\end{flalign}
where $\alpha_{KL} = \frac{1}{2\boldsymbol{h}^\top\boldsymbol{H}^{-1}\boldsymbol{h}}$, $\boldsymbol{h}$ is the gradient of the utility advantage function, $\boldsymbol{H}$ is the Hessian matrix of the KL divergence constraint, $M^\prime \leq \infty$ is some constant.
\end{lemma}
\begin{proof}
    Define the following set:
    \begin{flalign}
    Z_{\pi_k} = \big\{\pi\big| V_{c, {p^{c, \boldsymbol{\xi}}_k}}^{\pi_k}(\rho) + \mathbb{E}_{s\sim d^{\pi_k}_{p^{c, \boldsymbol{\xi}}_k}, a\sim \pi}[A_{c, {p^{c, \boldsymbol{\xi}}_k}}^{\pi_k}(s, a)] \geq V_{c, {p^{c, \boldsymbol{\xi}}_k}}^{\pi_k}(\rho)\big\}.
    \end{flalign}
    Note that the current policy $\pi_k$ lies in $Z_{\pi_k}$. Define the policy $\pi_{k+1}^l$ as the projection of $\pi_{k+\frac{1}{2}}$ onto $Z_{\pi_k}$. We have that 
    \begin{flalign}\label{eq:kl_vio}
    D_{KL}(\pi_{k+1}\|\pi_k)(s) = D_{KL}(\pi_{k+1}^l\| \pi_{k})(s)  + D_{KL}(\pi_{k+1}\|\pi_{k+1}^l)(s) + \big(\pi_{k+1}(s) - \pi_{k+1}^l(s)\big)^\top\log\frac{\pi_{k+1}^l(s)}{\pi_k(s)}.
    \end{flalign}
    From \ref{lemma:kl}, we have that $\mathbb{E}_{s\sim d^{\pi_k}_{p^{r, \boldsymbol{\xi}}_k}}[D_{KL}(\pi_{k+1}^l\| \pi_{k})(s)]\leq \delta$. For small $b$, $\mathbb{E}_{s\sim d^{\pi_k}_{p^{r, \boldsymbol{\xi}}_k}}[D_{KL}(\pi_{k+1}\|\pi_{k+1}^l)(s)]$ can be approximated by the second order expansion. We have that
    \begin{flalign}\label{eq:alpha}
    \mathbb{E}_{s\sim d^{\pi_k}_{p^{r, \boldsymbol{\xi}}_k}}[D_{KL}(\pi_{k+1}\|\pi_{k+1}^l)(s)] &\approx \frac{1}{2}(\boldsymbol{\theta}_{k+1} - \boldsymbol{\theta}^l_{k+1})^\top \boldsymbol{H}(\boldsymbol{\theta}_{k+1} - \boldsymbol{\theta}^l_{k+1})\nn\\& = \frac{1}{2}\Big(\frac{b}{\boldsymbol{h}^\top\boldsymbol{H}^{-1}\boldsymbol{h}} \boldsymbol{H}^{-1}\boldsymbol{h}\Big)^\top\boldsymbol{H} \Big(\frac{b}{\boldsymbol{h}^\top\boldsymbol{H}^{-1}\boldsymbol{h}} \boldsymbol{H}^{-1}\boldsymbol{h}\Big)\nn\\&= \frac{b^2}{2\boldsymbol{h}^\top\boldsymbol{H}^{-1}\boldsymbol{h}}\nn\\& = b^2 \alpha_{KL},
    \end{flalign}
    where $\alpha_{KL} = \frac{1}{2\boldsymbol{h}^\top\boldsymbol{H}^{-1}\boldsymbol{h}}$ and the first equality is from the update rule in \eqref{eq:update_approx}. 
    For $\big(\pi_{k+1}(s) - \pi_{k+1}^l(s)\big)^\top\log\frac{\pi_{k+1}^l(s)}{\pi_k(s)}$, we have that 
    \begin{flalign}
    \big(\pi_{k+1}(s) - \pi_{k+1}^l(s)\big)^\top\log\frac{\pi_{k+1}^l(s)}{\pi_k(s)} \leq \|\pi_{k+1}(s) - \pi_{k+1}^l(s)\|_1\big\|\log\frac{\pi_{k+1}^l(s)}{\pi_k(s)}\big\|_\infty.
    \end{flalign}
    Since $\mathbb{E}_{s\sim d^{\pi_k}_{p^{r, \boldsymbol{\xi}}_k}}[D_{KL}(\pi_{k+1}^l\| \pi_{k})(s)]\leq \delta$, there exists $M^\prime$ such that $\mathbb{E}_{s\sim d^{\pi_k}_{p^{r, \boldsymbol{\xi}}_k}}\Big[\big\|\log\frac{\pi_{k+1}^l(s)}{\pi_k(s)}\big\|_\infty\Big] \leq M^\prime.$ 
    Moreover, we have that $\|\pi_{k+1}(s) - \pi_{k+1}^l(s)\|_1 \leq \sqrt{2D_{KL}(\pi_{k+1}\|\pi_{k+1}^l)(s)}$.
    We then have that 
    \begin{flalign}\label{eq:beta}
    &\mathbb{E}_{s\sim d^{\pi_k}_{p^{r, \boldsymbol{\xi}}_k}}\Big[\big(\pi_{k+1}(s) - \pi_{k+1}^l(s)\big)^\top\log\frac{\pi_{k+1}^l(s)}{\pi_k(s)}\Big]\nn\\& \leq \mathbb{E}_{s\sim d^{\pi_k}_{p^{r, \boldsymbol{\xi}}_k}}\Big[M^\prime \sqrt{\frac{1}{2}D_{KL}(\pi_{k+1}\|\pi_{k+1}^l)(s)}\Big]\nn\\&\overset{(a)}{\leq} M^\prime \sqrt{\mathbb{E}_{s\sim d^{\pi_k}_{p^{r, \boldsymbol{\xi}}_k}}\Big[\frac{1}{2}D_{KL}(\pi_{k+1}\|\pi_{k+1}^l)(s)\Big]}\nn\\&\overset{(b)}{\approx} bM^\prime \sqrt{\frac{\alpha_{KL}}{2}},
    \end{flalign}
    where $(a)$ is from Jensen's inequality and $(b)$ is from \eqref{eq:alpha}.
    
    By combining \eqref{eq:kl_vio}, \eqref{eq:alpha} and \eqref{eq:beta}, we have that
    \begin{flalign}
    \mathbb{E}_{s\sim d^{\pi_k}_{p^{r, \boldsymbol{\xi}}_k}}[D_{KL}(\pi_{k+1}||\pi_k)(s)] \leq \delta + b^2\alpha_{KL} + bM^\prime\sqrt{\frac{\alpha_{KL}}{2}}.
    \end{flalign}
\end{proof}

With Lemma \ref{lemma:vio}, Theorem \ref{theorem:non-feasible} can be proved similarly as Theorem \ref{theorem:feasible}.

\section{Experiments}\label{sec:appendix_sim}
The detailed environments descriptions are in the following:

\textbf{Gambler Problem} in a game in which a gambler bets on a sequence of coin tosses, wining the stake when the outcome is head and losing when it's tail. Starting from an initial balance, the game ends once the gambler's balance reaches $16$ or $0$. For different state-action pairs, the gambler receives different utilities. The reward is $10$ when the balance reaches $16$ and $0$ otherwise. The probability of head for each coin toss is $p=0.6$. The radius of the uncertainty set is $0.1$. 

\textbf{$N$-chain problem} involves a chain with $N$ nodes. At each node, the agent can choose to move to its left or its right. Upon moving to its left, it receives a reward-utility signal of $(1, 0)$, while moving to its right yields a reward-utility signal of $(0, 2)$. When reaching the $N$-th node, the agent receives a bonus reward of $10$. With probability $0.1$, the agent may slip to the different direction of its action. We let $N = 40$ and the radius of the uncertainty set be $0.15$. 

\textbf{Frozen-Lake problem} is about training an agent to cross a $4\times 4$ frozen lake from the starting point to the end point without falling into any holes. Upon falling into the holes, the agent will get trapped and receive zero reward and utility. Reaching the end point yields a reward of $r=200$, otherwise $r=0$. At some states, the agent will receive a utility of $c=1$. The agent may slip to the different direction of its action. The radius of the uncertainty set is $0.1$.

\textbf{Point Gather} is a benchmark Mujoco task for constrained MDP, in which an agent is rewarded for gathering green apples but is constrained to collect a limited number of red fruit \cite{achiam2017constrained}.

\end{document}